\documentclass[jou,apacite,floatsintext]{apa6}
\usepackage{pdfpages}
\usepackage{CJKutf8,bbm}
\usepackage{tipa}
\usepackage{amsmath}
\usepackage[percent]{overpic}
\usepackage{subcaption}
\usepackage{graphicx}
\usepackage{subfiles}
\usepackage{csquotes}
\usepackage{etoolbox}
    \patchcmd{\maketitle}{\vspace*{1in}}{}{}{}
\usepackage{color}

\let\OldMaketitle\maketitle
\let\OldNewpage\newpage
\renewcommand\maketitle{%
   \renewcommand\newpage{\let\newpage\OldNewpage}
   \OldMaketitle%
   }
   
\usepackage{draftwatermark}
\SetWatermarkText{DRAFT}
\SetWatermarkScale{1}
\vspace{-1cm}
\title{Are words easier to learn from infant- than adult-directed speech? \\ A quantitative corpus-based investigation}

\shorttitle{the IDS lexicon}

\author{ 
{\large \bf Adriana Guevara-Rukoz$^1$,}
{\large \bf Alejandrina Cristia$^1$,}\\
{\large \bf Bogdan Ludusan$^1$,} 
{\large \bf Roland Thiollière$^1$,}
{\large \bf Andrew Martin$^2$,}\\
{\large \bf Reiko Mazuka$^{3,4}$,}
{\large \bf Emmanuel Dupoux$^1$\\}
\vspace{0.1cm}
$^1$Laboratoire de Sciences Cognitives et Psycholinguistique, ENS/EHESS/CNRS/PSL \\
$^2$Department of English Literature and Language, Faculty of Letters, Konan University \\
$^3$Laboratory for Language Development, RIKEN Brain Science Institute \\
$^4$Department of Psychology and Neuroscience, Duke University 
}

\affiliation{}
\abstract{
\small{We investigate whether infant-directed speech (IDS) could facilitate word form learning when compared to adult-directed speech (ADS). To study this, we examine the distribution of word forms at two levels, acoustic and phonological, using a large database of spontaneous speech in Japanese. At the acoustic level we show that, as has been documented before for phonemes, the realizations of words are more variable and less discriminable in IDS than in ADS. At the phonological level, we find an effect in the opposite direction: the IDS lexicon contains more distinctive words (such as onomatopoeias) than the ADS counterpart. 
Combining the acoustic and phonological metrics together in a global discriminability score reveals that the bigger separation of lexical categories in the phonological space does not compensate for the opposite effect observed at the acoustic level. As a result, IDS word forms are still globally less discriminable than ADS word forms, even though the effect is numerically small. We discuss the implication of these findings for the view that the functional role of IDS is to improve language learnability.
}}


\rightheader{Are words in infant-directed speech easier to learn?}
\leftheader{Guevara-Rukoz \textit{et al.}}

\begin{document}
\maketitle

\setcounter{secnumdepth}{3}

\section{Introduction}
Infants' language acquisition proceeds at an amazing speed despite the inherent  difficulties in discovering linguistic units such as phonemes and words from continuous speech. A popular view holds that part of the problem may be alleviated by the infants' caregivers, who may simplify the learning task when they speak to their infants in a particular register called infant-directed speech (IDS). In this paper, we compare IDS and adult-directed speech (ADS) in terms of dimensions that are relevant for the learnability of sound categories. We first review alternative hypotheses about a possible facilitatory role of IDS.

\subsection{IDS-ADS differences in the context of learnability}

The notion that particular speech registers may have articulatory and acoustic properties that enhance speech perception may have been first introduced by Lindblom in the context of his Hyper and Hypo-articulation (H\&H) theory \citeyear{lindblom1990}. In the case of hyperarticulation, the resulting listener-oriented modifications are referred to as `hyperspeech'. Here, the priority is to enhance differences among contrasting elements, and it runs counter the speaker-oriented tendency to produce more economical articulatory sequences. 

\citeA{fernald2000} proposed a more general definition of hyperspeech in the context of language acquisition. The idea is that parents may manipulate linguistic levels other than articulatory ones, such as information relating to word frequency or neighborhood density, resulting in facilitated perception: 

\begin{displayquote}
``[T]he hyperspeech notion should not be confined to articulatory factors at the segmental level, but should be extended to a wider range of factors in speech that facilitate comprehension by the infant''. 
\end{displayquote}

While the hyperspeech notion initially refers to a modification of language as to enhance perception, \citeA{kuhl1997} go one step further, positing that IDS register-specific modifications may also enhance \textit{learning}:

\begin{displayquote}
``Our findings demonstrate that language input to infants has culturally universal characteristics designed to promote language learning''.
\end{displayquote}

We call this last hypothesis the \textit{Hyper Learnability} Hypothesis (HLH). It goes beyond the hyperspeech hypothesis in that it refers not to perception but to the language learning processes operating in the infant.  Importantly, these two notions may not necessarily be aligned.
In some instances, both hyperspeech and HLH are congruent with the usually reported properties of IDS: exaggerated prosody and articulation \cite{fernald1989, soderstrom2007}, shorter sentences \cite{phillips1973, newport1977, fernald1989}, simpler syntax \cite{phillips1973, newport1977}, and slower speech rate \cite{fernald1989, englund2005} (see \citeNP{soderstrom2007, golinkoff2015} for more comprehensive reviews). All of these properties are plausible candidates for facilitating both language perception and language learning at the relevant linguistic levels — namely phonetic, prosodic, lexical and syntactic — by making these features more salient or more contrastive to the infant. Yet, in other instances, perception and learning may diverge. As \citeA{kuhl_2000} notes:

\begin{displayquote}
``Mothers addressing infants also increase the variety of exemplars they use, behaving in a way that makes mothers resemble many different talkers, a feature shown to assist category learning in second-language learners.''
\end{displayquote}

In this case, increase in variability, which is known to negatively affect speech perception in both adults and children (see \citeNP{mullennix_1989,ryalls_1997,bergmann2016}) is nevertheless hypothesized to positively affect learning in infants. Work by \citeA{rost2009} suggests that this might be the case for 14-month-old infants learning novel word-object mappings. However, it appears that not any kind of variability will do; only increased variability in certain cues --specifically those irrelevant to the contrasts of interest-- promoted learning of word-object mappings \cite{rost2010}.  
This illustrates the very important point that HLH cannot be empirically tested independently of a specific hypothesis or theory of the learning process in infants. 
Ideally, the hypothesis or theory should be explicit enough that it could be implemented as an algorithm, which derives numerical predictions on learning outcomes when run on speech corpora of ADS and IDS \cite{dupoux_2017}. Unfortunately, as of today, such algorithms are not yet available for modelling early language acquisition in infants. Yet, a reasonable alternative is to resort to measurements that act as a \textit{proxy} for learning outcomes within a given theory.

In the following, we focus on a component of language processing which has been particularly well studied: speech categories. For this component, a variety of theories have been proposed, which can be separated in two types: bottom-up theories and top-down theories. We review these two types in the following sections and discuss possible proxies for them.

\subsection{Bottom-up theories: Discriminability as a proxy}

Bottom-up theories propose that phonetic categories emerge from the speech signal; they are extracted by attending to certain phonetic dimensions \cite{jusczyk_1990}, or by identifying category prototypes \cite{kuhl_1993}. More explicitly, \citeA{maye_2002} proposed that infants construct categories by tracking statistical modes in phonetic space. This idea can be made even more computationally explicit by using unsupervised clustering algorithms, such as Gaussian mixture estimation (\citeNP{deboer2003,vallabha_2007,mcmurray_2009,lake_2009}), or self-organizing neural maps (\citeNP{kohonen_neural_1988,guenther_1996,vallabha_2007}). Given the existence of such computational algorithms, it would seem easy to test if IDS enhances learning by running them on IDS and ADS data, and then evaluating the quality of the resulting clusters. 

However, this is not so simple for two reasons. First, each of the above-mentioned algorithms makes different assumptions about the number, granularity, and shape of phonetic categories, parameters which could potentially lead to different outcomes. Even more problematic is that this subset of algorithms does not exhaust the space of possible clustering algorithms. Since we do not know which of these assumptions and algorithms are those that best approximate computational mechanisms used by infants, applying these algorithms to data may not get us any closer to a definitive answer.
Second, these particular algorithms have only been validated on artificially simplified data (e.g., representing categories as formant measurements extracted from hand-segmented data) and not on a corpus of realistic speech. In fact, when similar algorithms are run on real speech, they fail to learn phonetic categories; instead they learn smaller and more context-dependent units (e.g., \citeNP{varadarajan_2008}; see also \citeNP{antetomaso_2016}). The unsupervised discovery of phonetic units is currently an unsolved problem which gives rise to a variety of approaches (see \citeNP{versteegh_2016} for a review).

Given the unavailability of effective phoneme discovery algorithms that could test the bottom-up version of HLH, many researchers have adopted a more indirect approach using descriptive measures of phonetic category distributions as a \textit{proxy} for learnability. Here we review two such proxies: category \textit{separation} and category \textit{discriminability}.

Category separation corresponds to the distance between the center of these categories in phonetic space. \citeA{kuhl1997} measured the center of the `point' vowels \textipa{/a/}, \textipa{/i/}, and \textipa{/u/} in formant space, in ADS and IDS, across three languages (American English, Russian and Swedish). Results revealed that the spatial separation between the center of these vowels was increased in IDS compared to ADS. 
This observation has been replicated in several studies (\citeNP{bernsteinr1984, andruski1999, burnham2002, liu2003, uther2007, cristia2014, mcmurray2013}, although see \citeNP{benders2013}). However, it is less clear that separation generalizes to other segments beyond the three point vowels. For instance, \citeA{cristia2014} attested increased separation of the point vowels in speech spoken to 4- and 11-month old learners of American English, but not for other vowel contrasts (\textit{e.g.,} \textipa{[i-\textipa{I}]}). The between-category distance among the latter vowel categories was not larger in IDS than in ADS (see also \citeNP{mcmurray2013} for similar results). This is problematic for learnability because one might argue on computational grounds that the vowels that are difficult to learn are probably not the point vowels which are situated at the extreme of the vocal space, but rather the ones that are in the middle and have several competitors with which they can be confused. 

\begin{figure}[ht!] 
\centering
    \vspace{0.5cm}
    \begin{overpic}[width=\linewidth]{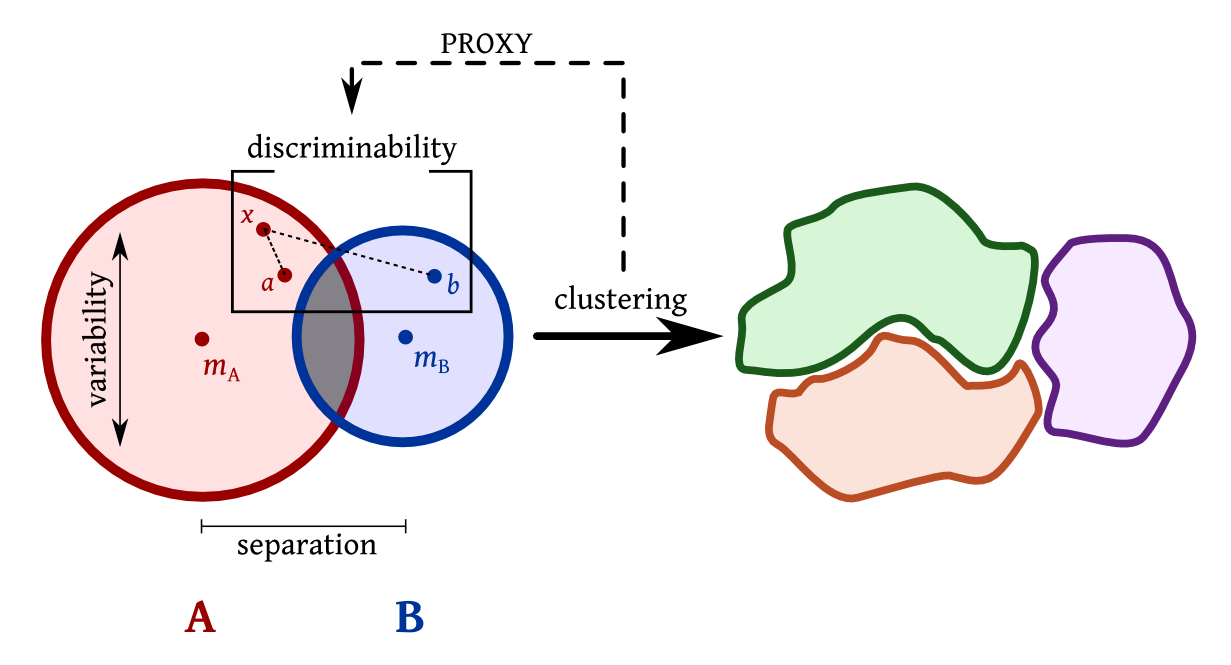}
    \end{overpic}
    \vspace{0.5cm}
    \caption{Schematic view of separation, variability and discriminability between two categories A and B (left), and a possible clustering obtained from the distributions (right).
    \emph{Separation} measures the distance between the center of categories A and B; it is computed as the distance between the medoids $\textit{m}_{A}$ and $\textit{m}_{B}$.
    \emph{Variability} measures the spatial spread of tokens within a given category; it is computed as the average distance between tokens in a category. 
    \emph{Discriminability} depends on both variability and separation; it is quantified with an ABX score as the probability that a given token $\textit{x}$ (say, of A) is less distant to another token $\textit{a}$ of A than to a token $\textit{b}$ of B.} 
    \label{fig:schema_dist}

\end{figure}

There is another reason to doubt that separation is a very good proxy in the first place. As shown in Figure \ref{fig:schema_dist}, categories are defined not only by their center, but also by their variability. 
If, for instance, IDS not only increases the separation between category centers compared to ADS, but also increases within category variability, the two effects could cancel each other out or even wind up making IDS more difficult to learn.
In fact, as we mentioned above, \citeA{kuhl1997} reported that parents tend to be more variable in their vowel productions in IDS than ADS. This was confirmed in later studies \cite{kirchhoff2005,mcmurray2013,cristia2014}. If so, what is the net effect of these two opposing tendencies on category learnability? 

Previous work by \citeA{schatz_2016} has shown that the performance of unsupervised clustering algorithms can be predicted by a psychophysically-inspired measure: the ABX \emph{discrimination score}. The intuition behind this measure is illustrated in Fig. \ref{fig:schema_dist}: it is defined as the probability that tokens within a category are closer to one another than between categories. If the two categories are completely overlapping, the ABX score is 0.5. If, on the other hand, the two categories are well segregated, the score can reach 1\footnote{\citeA{schatz_2016} has shown that an ABX score of 1 between categories A and B implies that the two categories can be discovered without error by the clustering algorithm \textit{k}-means.}. This work has demonstrated that the ABX score tends to be more statistically stable than standard clustering algorithms (k-nearest neighbors, spectral clustering, hierarchical clustering, k-means, etc.) while predicting their outcomes better than they predict each other's outcomes. All in all, this method is independent of specific learning algorithms, is non-parametric (i.e. it does not assume particular shapes of distributions) and can operate on any featural representation including raw acoustic features. It can therefore be used as a stable proxy of unsupervised clustering and, therefore, of bottom-up learnability. 


Using this measure, \citeA{martin2015} systematically studied the discriminability of 46 phonemic contrasts of Japanese by running the ABX discriminability test on a speech corpus with features derived from an auditory model, namely mel spectral features. The outcome was that, on average, phonemic categories were actually \emph{less discriminable} in IDS than in ADS. While most contrasts did not differ between the two registers, the few that systematically differed pointed rather towards a decrease in acoustic contrastiveness in IDS at the phonemic level.

To sum up, if one uses ABX-discriminability as a proxy for bottom-up learnability, we can conclude that the HLH is not supported by the data available.  However, bottom-up learning is not the only theoretical option available to account for phonetic learning in infants. Next, we examine top-down theories.

\subsection{Top-down theories: Three learnability subproblems}

\begin{figure}[ht!] 
\centering
    \vspace{0.5cm}
    \begin{overpic}[width=\linewidth]{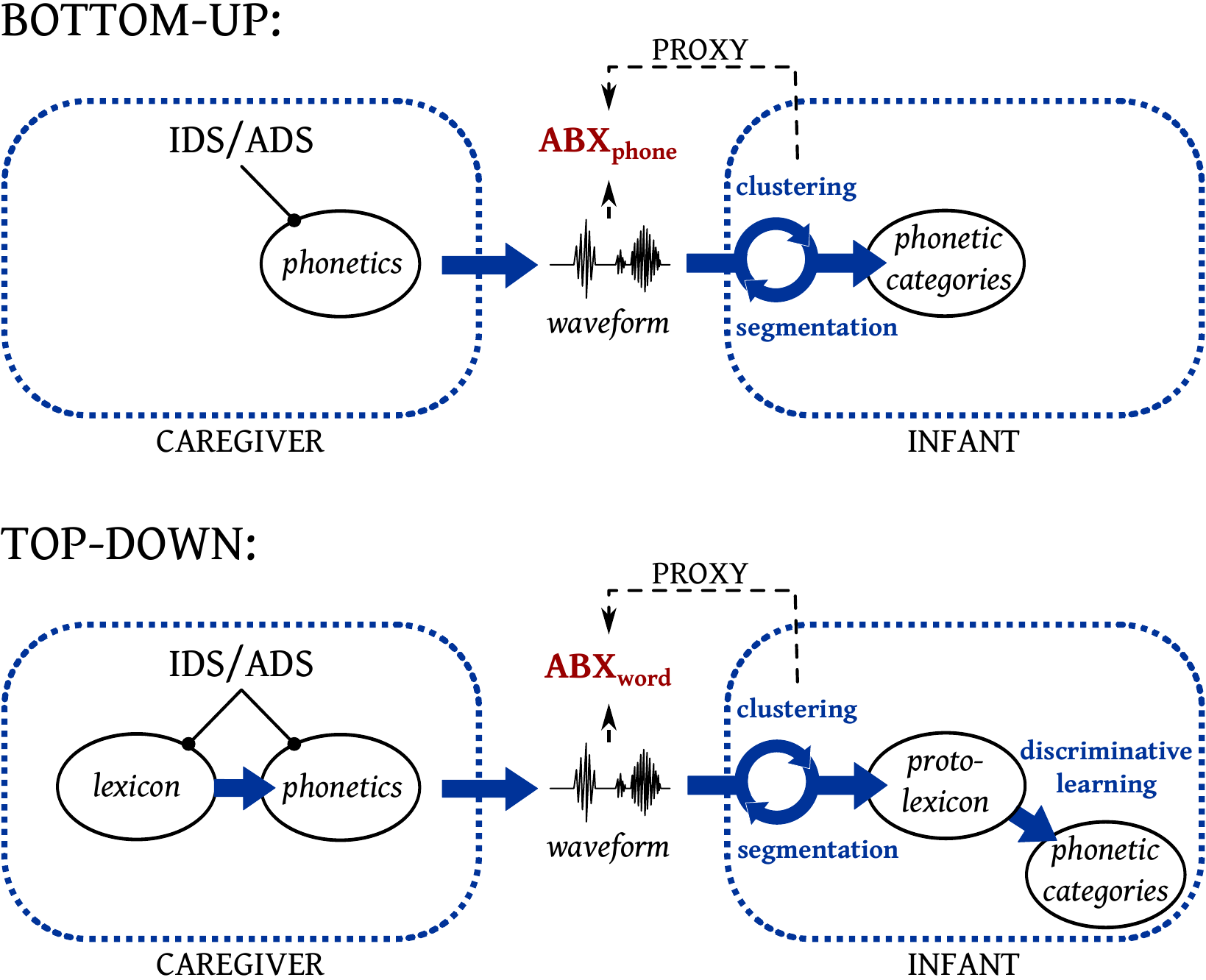} 
    \end{overpic}
    \hspace{0.5px}
    \caption{Schematic view of (a) bottom-up  and (b) top-down models of phonetic learning, together with ABX discriminability as a proxy for measuring the effect of ADS versus IDS on learnability.} 
    \label{fig:schema_models}

\end{figure}

Top-down theories of phonetic category learning share with linguists the intuition that phonemes are defined, not so much through their acoustic properties, but rather through their function. The function of phonemes is to carry meaning contrasts at the lexical level. Top-down theories therefore posit that phonemes emerge from the lexicon. As stated by \citeA{werker_2005} (see also \citeNP{beckman_2000}):
\begin{displayquote}
``As the vocabulary expands and more words with overlapping features are added, higher order regularities emerge from the multidimensional clusters. These higher order regularities gradually coalesce into a system of contrastive phonemes.'' (p. 217)
\end{displayquote}


There are many ways to flesh out these ideas in terms of computational mechanisms. All of them involve at least the requirement that (some) word forms are learned and that these forms constrain the acquisition of phonetic categories. This can be summarized in terms of three subproblems (Fig. \ref{fig:schema_models}b): (1) segmenting word tokens from continuous speech, (2) clustering said word tokens into types, (3) using said types to learn phonetic categories via a contrastive mechanism. 
Arguably, these three subproblems are interdependent (in fact, some models address several of them jointly, e.g., \citeNP{feldman_learning_2009} or iteratively, e.g., \citeNP{versteegh_2016}), and only a fully specified model would enable to fully test the functional impact of IDS for learnability under such a theory. Yet, as above, we claim that one can develop measures that can act as proxies for learnability, even in the absence of a full model. 

In what follows, we focus on the second subproblem, i.e., the clustering of word types, which we take to be of central importance for phonetic category learning. Indeed, in case of a failure to solve subproblem 1 (e.g., infants undersegment ``\textit{the dog}'' into ``\textit{thedog}'', or oversegment ``\textit{butterfly}'' into ``\textit{butter fly}''), it is still possible to use contrastive learning with badly segmented proto-words to learn phonetic categories \cite{fourtassi2014}. In contrast, in case of a failure to solve subproblem 2, (e.g., infants merge ``\textit{cat}'' and ``\textit{dog}'' into a signal word type, or split ``\textit{tomato}'' into many context or speaker dependant variants) then it is much more dubious that contrastive learning can be of any help to establish phonetic categories. Our experiments therefore only address subproblem 2, and we come back to the other two subproblems in the General Discussion.

\subsection{The present study: word form discriminability}

The construction of word form categories is a similar computational problem to the problem of constructing phonetic categories discussed above. Both can be formulated as unsupervised clustering problems, the only difference being the granularity and number of categories being formed. Instead of sorting out instances of `i', `a' and `o' into clusters, the problem is to sort out instances of `cat', `dog' and `tomato' into clusters. Therefore in both instances, it is possible to use ABX discriminability as a proxy for the (bottom-up) learnability of these categories. Of course, words being composed of phonemes, one would expect a correlation between ABX discriminability on phonemes and on words. However, the word form level introduces two specific types of effects making such a correlation far from trivially true.

First, the word level typically introduces specific patterns of phonetic variability. For instance, the word `tomato' can be produced in a variety of ways: /\textipa{t\super h@"meIRoU}/, /\textipa{t@"meIt@}/, /\textipa{t@"mAːt@U}/, etc. Some of these variations are dependant on the dialect but others can surface freely within speaker, or depending on context, speaking style, or speaking rate. Such phonetic effects translate into distinct acoustic realizations of the word forms, potentially complicating the task of word form category learning. Could it be that IDS limits this source of variation, thereby helping infants to construct word form categories? Some studies have shown the use of more canonical forms in IDS than ADS (e.g., \citeNP{dilley_2014}), while others have not (e.g., \citeNP{fais_2010,lahey_2014}), but to our knowledge no study has looked at the global effect of these variations on word discriminability, and done so systematically. This is what we will examine in Experiment 1.

Second, and setting aside phonetic realization to focus on abstract phonological characteristics, words tend to occupy sparse regions of phonological space. Put differently, there are many more unused possible word forms than actual ones. This results in minimal pairs being generally rare. For instance, a corpus analysis reveals that, in English, Dutch, French, and German, minimal pairs will concern less than 0.1\% of all pairs \cite{dautriche2017}; in fact, two words selected at random will differ in more than 90\% of their phonemes on average.
This should make word form clustering an easier task than phonetic clustering, a welcome result for top-down theories. However, it could be that IDS modulates this effect by containing a different set of words than the vocabulary directed to adults. Corpora descriptions of IDS suggest that this is the case: Caregivers use a reduced vocabulary \cite{phillips1973, kaye1980, henning2005}, which often includes a set of lexical items with special characteristics, such as syllabic reduplications and mimetics \cite{ferguson1964, fernald1993, mazuka2008}. May IDS boost learning by containing more phonologically distinct word forms than ADS? This is what we will examine in Experiment 2. 


The overall learnability of word forms, as far as clustering is concerned, is the combined effect of phonetic/acoustic discriminability (isolated in Experiment 1) and phonological discriminability (isolated in Experiment 2). As these two factors may go in different directions, we study the global discriminability of IDS versus ADS word form lexicons in Experiment 3.




\subsection{Japanese IDS} 

Like other variants of IDS around the globe \cite{ferguson1964}, Japanese IDS is characterized by the presence of Infant-Directed Vocabulary (IDV), `babytalk' specifically used when interacting with infants. According to a survey and corpora studies by \citeA{mazuka2008}, these words are mostly phonologically unrelated to words in the ADS lexicon. In particular, IDV presents many instances of reduplications (around 65\%) and onomatopoeias/mimetic words (around 40\%).\footnote{
In a study by \citeA{fernald1993}, Japanese mothers used onomatopoeitic words more readily than American mothers.} 
Phonological structures found in IDV are, in fact, more similar to phonological patterns produced by Japanese infants earlier in development than to patterns found in the adult lexicon (\citeNP{tsuji2014segmental}; a list of 50 earlier produced words is given by \citeNP{iba2000}). In addition to pattern repetition within words, IDS also presents more content word repetition, as well as more frequent and longer pauses, making utterances in IDS shorter than in ADS \cite{martin2016}.       

Regarding the phonetics of Japanese IDS, it presents pitch-range expansion \cite{igarashi2013}, but it is not slower than ADS when taking into account local speech rate \cite{martin2016}. More related to our question of phonetic categories, vowel space expansion in F1 x F2 space has been attested in Japanese IDS \cite{andruski1999, miyazawa2017}; however, IDS categories presented higher variability and overlap \cite{miyazawa2017}, consistent with the decrease of acoustic discriminability observed by \citeA{martin2015}. In fact, contrary to intuition, IDS appears to present more devoicing of non-high vowels than ADS (i.e., less canonical and identifiable tokens), due to breathiness \cite{martin2014}. This paralinguistic modification of speech, which is thought to convey affect, is more prevalent in IDS than ADS \cite{miyazawa2017}. 

\subsection{Corpus}

Most of the Japanese studies cited above, as well as the work described in this paper, have used data from the RIKEN Japanese Mother-Infant Conversation Corpus, R-JMICC \cite{mazuka2006}, a corpus of spoken Japanese produced by 22 mothers in two listener-dependent registers: infant-directed speech (IDS) and adult-directed speech (ADS) \cite{igarashi2013}. 

For our study, a word was defined as a set of co-occurring phonemes with word boundaries following the gold standard for words in Japanese, roughly corresponding to dictionary entries. Lexical derivations were considered to belong to a separate type category with respect to their corresponding lemmas. For instance, \emph{/nai/} and \emph{/aru/}, inflections of the verb \begin{CJK}{UTF8}{goth}ある\end{CJK} \emph{/aru/} (English: \textit{to be}), were evaluated as separate words. Homophones were collapsed into the same word category in the analyses. 

Because of the emphasis given to phonological structure when defining word categories, devoiced vowels were considered to be phonologically identical to their voiced counterparts, and similarly for abnormally elongated vowels or consonants that did not result in lexical modifications (i.e. use of gemination for emphasis). Additionally, fragmented, mispronounced, and unintelligible words were not included in our analyses (approximately $5,34\%$ out of the initial corpus). The resulting corpus is henceforth referred to as the \emph{base corpus}; information about its content can be found in Table \ref{table:basecorpus}.  

\begin{table}
\centering
\caption{Description of the base corpora for ADS and IDS}
\vspace{0.5cm}
\label{table:basecorpus}
\begin{tabular}{lccc}
    \toprule
    & ADS   & IDS    &  \\
    \midrule
Duration & 3 hours  & 11 hours  &  \\
Types     & $1,382$     & $1,765$  &  \\
Tokens    & $12,248$    & $34,253$  & \\ 
    \bottomrule
\end{tabular}
\end{table}


\section{Experiment 1: Acoustic distribution of word tokens}

In this experiment, we ask whether caregivers articulate words in a more or less `distinctive' manner when addressing their infants. Our aim is to answer this question at a purely acoustic level, \textit{i.e.}, taking into account phonetic and acoustic variability, after removing influences from other aspects that vary across registers (e.g., lexical structure). Therefore, the following analyses have been restricted to the lexicon of words that are \emph{common} to IDS and ADS for each parent. 

Our main measure is ABX discriminability applied to entire words. As in \citeA{martin2015}, we use the $ABX_{score}$ which shows classification at chance with a value of 0.5, while perfect discrimination yields a score of 1. As such, a higher $ABX_{score}$ for IDS than ADS would mean that, on average, parents make their word categories more acoustically discriminable when addressing their infants, making these words easier to learn according to top-down theories.

The ABX discriminability measure implies computing the acoustic distance between word tokens, and computing the probability that two tokens belonging to the same word type are closer to one another than two tokens belonging to two distinct word types. 

Since it is the first time that such a discriminability measure is used at the word level, we validate it in a control condition in which there are \textit{a priori} reasons to expect differences in discriminability between two speech registers. Namely, we assess the discrimination of words common to ADS and read speech (RS). This register is typically articulated in a slower, clearer, and more canonical fashion than spontaneous speech. Knowing this, we expect the $ABX_{score}$ to be higher in read speech (RS) than in spontaneous speech (ADS).

Moreover, in order to further validate the application of our method to word units, two additional sub-measures are explored, following the distinctions introduced in Fig. \ref{fig:schema_dist}: between-category \emph{separation} and within-category \emph{variability}. 

\subsection{Methods} 
\subsubsection{Control corpus}
The Read Speech (RS) subsection of the RIKEN corpus consists of recordings from a subset of 20 out of the 22 parents which had also previously been recorded in the ADS and IDS registers. Participants read 115 sentences containing phonemes in frequencies similar to those of typical adult-directed speech \cite{sagisaka1990}. We extracted the words that were common to the read and the ADS subcorpora for each individual parent. We obtained between 19 and 32 words, each of them having between 2 and 49 occurrences. All of these word tokens were selected for subsequent analysis in the control ADS \textit{vs.} RS comparison. 

\subsubsection{Experimental corpus}
All 22 participants had data in the IDS and ADS registers. For each participant, we selected the words that were common to the two registers. We obtained between 43 and 64 word types (individual numbers can be seen in the Appendix Table A1). All of the word tokens for these types were selected for subsequent analyses in the experimental condition comparing ADS vs. IDS. We did not match IDS and ADS on number of tokens per type to maximize the reliability of the metrics. Since ABX is an unbiased metric of discriminability, the size of a corpus will only modulate the standard error, not the average of the metric. It therefore cannot bias the discriminability score in IDS vs. ADS; simply the fact that the ADS scores are estimated from a smaller corpus means that they will be noisier than the IDS scores. Matching the IDS corpus size to that of ADS would result in increasing the noise in the IDS scores. Number of total tokens per speaker are shown in the Appendix Fig. A1. 

\subsubsection{Acoustic distance}
The three acoustic measures that were computed, namely separation, variability, and discriminability ($ABX_{score}$), all depend on a common core function which provides the measure of acoustic distance between two word tokens. 

As in \citeA{martin2015}, we represented word tokens using compressed mel filterbanks, which corresponds to the first stage of an auditory model \cite{moore1997, schatz_2016}.

Specifically, the audio file of each token was converted into a sequence of auditory spectral frames sampled 100 times per second, obtained by running speech through a bank of 13 band-pass filters centered on frequencies spread according to a mel scale between 100 and 6855 Hz \cite{schatz_2013}. The energy of the output of each of the 13 filters was computed and their dynamic range was compressed by applying a cubic root. In summary, word tokens were represented as sequences of frames, which are vectors with 13-dimensions (i.e. 1 value per filter).  

The distance between a pair of tokens was computed as follows. First, the two tokens of interest were realigned in the time domain by performing dynamic time warping (DTW; \citeNP{sakoe1978}): this algorithm searches the optimal alignment path between the sequences of frames of the two tokens that are being compared. The distance between two aligned frames being compared was set to be the angle between the two 13-dimensional feature vectors representing said frames. Secondly, the average of the frame-wise distances along the optimal alignment path was set as the distance between that pair of tokens. 

Each of the three measures was computed separately for each speaker, both for IDS and for ADS. 

\subsubsection{Discriminability}

Discriminability calculations were performed as in \citeA{martin2015}, by estimating the probability that two tokens within a category are less distant than two tokens in two different categories. This score is computed for each pair of word types, and then aggregated by averaging across all of these pairs ($ABX_{score}$). 
The calculations were done using the ABXpy package available on \url{https://github.com/bootphon/ABXpy}.

More specifically, for each pair of word types A and B, we compiled the list of all possible (\emph{a,b,x}) triplets where \emph{a} was a token of category \emph{A}, \emph{b} a token of category \emph{B} and \emph{x} a token of either \emph{A} or \emph{B}. For instance, for word types \emph{$A = /nai/$} and \emph{$B = /aru/$}, there could be a triplet with tokens $a = [nai]_{1}$, $b = [aru]_{1}$, and $x = [nai]_{2}$. The distance \emph{$d(a,x)$} between tokens \emph{a} and \emph{x} was compared to the distance \emph{$d(b,x)$} between tokens \emph{b} and \emph{x}. In this example, since both \emph{a} and \emph{x} are tokens of category \emph{A}, we expect the acoustic distance between them to be smaller than their distance to a token belonging to a different category (i.e. token \emph{b} of type \emph{B}). 

As such, if $d(a,x) > d(b,x) $ (i.e. $[nai]_{2}$ more similar to $[aru]_{1}$ than to $[nai]_{1}$), the response given by the algorithm was deemed to be incorrect and an ABX$_{score}$ of $0$ was assigned to that specific triplet. On the other hand, if as expected $d(a,x) < d(b,x)$, the algorithm returned a response deemed as correct and a score of $1$ was given to the triplet. A final mean $ABX_{score}$ for all triplets was then computed for each speaker, separately for IDS and ADS, only taking into account word pairs that were observed in both speech registers. 

\subsubsection{Separation}

For each pair of word types, we computed the distance between their medoids. A medoid is defined as the word token which minimizes the average distance to all of the other tokens in that word type. In case of ties, we used a set of medoids, and their scores were averaged. Separation can be viewed as a generalization of the notion of phonetic expansion, except that it applies to entire word forms instead of particular segments (e.g., vowels).

\subsubsection{Variability}

For each word type, variability was computed as the average distance between each token and every other token within the same word type. By definition, only word types with more than one token were included in the calculation. One can view this measure as analogous to the standard deviation in univariate distributions.

\subsection{Results and discussion}
\begin{figure*}[ht!]
\centering
    \includegraphics[page=1, width=0.45\linewidth]{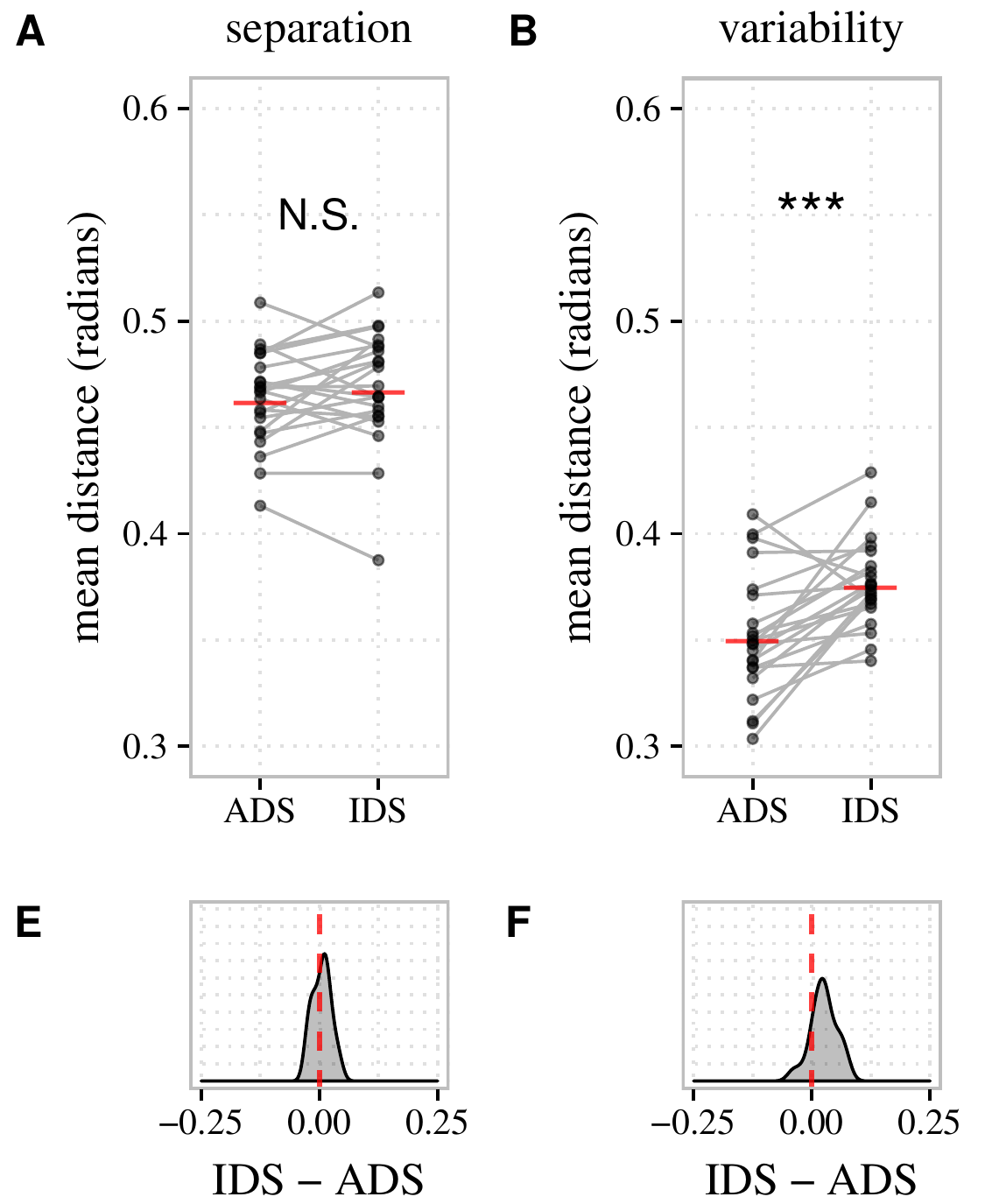}
    \hspace{0.75cm}
    \includegraphics[page=2, width=0.45\linewidth]{img/revision_figs.pdf}
    \vspace{0.5cm}
    
    
    \caption{Acoustic distinctiveness scores computed on word types common to IDS and ADS (panels A, B, C, E, F, G), or computed on word types common to ADS and RS (control condition; panels D and H). Upper panels display the distribution of the scores across speakers, as well as means within a speech register (red horizontal lines). Grey lines connect data points corresponding to the same caregiver in both registers (either ADS-IDS or ADS-RS). Bottom panels show the distribution of IDS minus ADS (or RS minus ADS) score differences. Densities to the right of the red zero line denote higher scores for IDS (or RS). 
    A, E: Mean between-category separation (ADS \textit{vs.} IDS).
    B, F: Mean within-category variability (ADS \textit{vs.} IDS).
    C, G: Mean ABX discrimination score (ADS \textit{vs.} IDS).
    D, H: Mean ABX discrimination score (ADS \textit{vs.} RS; control condition).
    $N.S.$: Non-significant difference. ***: $p < .001$. ****: $p < .0001$}
    \label{fig:exp1}
\end{figure*}

Regarding the control condition, we compared the acoustic discriminability of the word types common to ADS and RS. We obtained an average ABX discriminability score per speaker per register (ADS or RS). A paired Student's \textit{t-}test revealed that words were significantly more discriminable in RS than in ADS ($t(19) = 8.74; p < .0001$; \textit{Cohen's d} = $2.68$), with RS having an $ABX_{score}$ 0.09 points higher than ADS, on average ($ABX_{score}$ of $92\%$ versus $83\%$, respectively). As shown in Fig. \ref{fig:exp1} (panels D and H), all 20 parents showed this effect; individual scores can be found in the Appendix Table A1. In other words, on average the algorithm made twice as many errors classifying word tokens into categories in ADS compared to RS. This confirms that the ABX measure is able to capture the expected effects of read versus spontaneous speech on acoustic discriminability.

Focusing on the experimental condition, for each of the three measures (discriminability, separation, variability), we computed an aggregate score across word types separately for each parent and register (individual scores can be found in the Appendix Table A1). We then analysed the effect of register by running a paired Student's \textit{t}-test across parents.

The results are visually represented in Fig. \ref{fig:exp1}. First, the analysis revealed a numerically small but statistically reliable degradation in acoustic discriminability of words in IDS compared to ADS ($ABX_{score}$ IDS: $80\%$ \textit{vs.} ADS: $84\%$; $t(21) = -4.73; p < .001$; \textit{Cohen's d} = $-0.84$). This is consistent with the degradation in discriminability previously observed at the level of individual phonemes \cite{mcmurray2013, martin2015}.
Second, the trend for greater separation of word categories in IDS compared to ADS  was not statistically significant (IDS: $0.47$ rad \textit{vs.} ADS: $0.46$ rad; $t(21) = 1.23; p > .05$; \textit{Cohen's d} = $0.21$). 
Finally, there was a reliable increase in variability in IDS relative to ADS (IDS: $0.38$ rad \textit{vs.} ADS: $0.35$ rad; $t(21) = 4.28; p < .001$; \textit{Cohen's d} = $1.0$).  This increased variability is consistent with what has been observed at the level of individual phonemes \cite{mcmurray2013, cristia2014}.

In sum, we found that word discrimination is more easily achieved in ADS than in IDS. This can be analysed as being due to a large increase in variability in IDS which is not being compensated for by a necessary increase in separation. This is in contrast to predictions posited by the \textit{HLH}, but consistent with previous work at the phonemic level \cite{martin2015}. In a way, this is not a totally surprising result, since by virtue of matching word types across registers, the effect of register on phoneme variability and discriminability is passed on to the level of words. What is new, however, is that the IDS register does not compensate for the phonetic variability by producing more canonical word forms. Next, we examine the content of the lexicon in the two registers.

\section{Experiment 2: Phonological density} 

In this experiment, we focus on the phonological structure of the IDS and ADS lexicons. The core question is whether parents would select a set of words that are somewhat more `distinctive' in IDS, yielding a sparser lexicon. Such a sparse lexicon could compensate for the increased phonetic variability measured in Experiment 1, thereby helping infants to cluster word forms into types. 

We use Normalised Edit Distance (NED) as our main measure of the sparseness of the IDS and ADS lexicons. NED is defined as the proportion of changes (i.e., segmental additions, deletions and substitutions) to be performed in order to transform one word into another. The smaller the edit distance between two words, the more structurally similar they are.

NED takes into consideration not only phonological neighbors (i.e. words that differ by one phoneme), but also higher order neighbors when evaluating variation of the phonological structure of the lexicon in a psychologically relevant way. It is the direct phonological equivalent of the \emph{separation} metric used in Experiment 1. 
Indeed, both metrics measure the average distance between word categories: \textit{separation} measures acoustic distance, while \textit{NED} measures phonological distance. Experiment 1 showed that parents do not reliably expand the acoustic space when using IDS; Experiment 2 asks: Are they expanding the phonological space when using this register? 

Before moving on to the analysis, we point out that mean NED may vary with lexicon size. Indeed, as more and more words are added to a lexicon, changes in the neighborhood structure are to be expected. Typically, short words tend to have denser neighborhoods as the lexicon size increases (as the combinatorial possibilities for constructing distinct short words quickly saturate). At the same time, the ratio between short and long words tends to decrease with lexicon size, because most new additions in a lexicon tend to be long, and long words tend to have sparser neighborhoods than short words. In order to limit the influence of such properties on our results, IDS and ADS corpora were \textit{matched in lexicon size} before any comparison was performed. 

\subsection{Methods} 

\subsubsection{Sampling}

As can be seen in Table \ref{table:basecorpus}, the volume of data available for both speaking registers in the \emph{base corpus} was imbalanced; the IDS subset of the corpus contains more words (types and tokens) than its ADS counterpart. In order to account for this mismatch, we performed a frequency-dependent sampling of word types that matched their number in both speech registers. Types which were more frequently uttered by a speaker had a higher probability of being included in a sample than rarer ones. Moreover, since the measurement used in this section heavily relies on the nature of the words sampled, and as a way to increase estimation reliability, sampling was performed one hundred times per speaker per register.  For instance, if a speaker uttered 82 word types in ADS and 237 in IDS, we created 100 subsets of the IDS lexicon by sampling 82 types from the 237 available 100 times. The final metric for said speaker in a given speech register was the mean NED obtained from the corresponding 100 samples. On average, a sample contained $179.64 \pm49$ word types (see Table A2 of the Appendix for more information).

\subsubsection{Normalized Edit Distance}
For each parent, within each speech register, we computed the edit distance (ED) between every possible pair of types in the sampled lexicons. ED, also called the Levenstein distance, is defined as the minimal number of additions, deletions or substitutions needed to transform one string into another. It is computed using an algorithm very similar to the Dynamic Time Warping (DTW) algorithm used in Experiment 1; the algorithm finds a path that minimizes the total number of edits (insertions, deletions and substitutions, all of them equally weighted). The maximal number of changes $max(x,y)$ is defined as the maximum length of the two types $X$ and $Y$ under comparison. 
Normalized edit distances (NEDs) were therefore derived as follows:
\[NED_{XY} = \frac{ED_{XY}}{max(x,y)}\]
where \emph{x} and \emph{y} correspond to the phonemic lengths of two distinct words $X$ and $Y$. 
For instance, the ED between `tall' \textipa{/tOl/} and `ball' \textipa{/bOl/} is 1 (one substitution: \textipa{/t/} \(\Rightarrow{}\) \textipa{/b/}). Both words are 3 phonemes long, so $max(x,y) = 3$. Therefore, the NED between these types is \( \frac{1}{3}  \).
The more structurally similar two types are, the closer their NED will be to zero. 

\subsection{Results and Discussion}

\begin{figure}[ht!]
\centering
    \includegraphics[page=3, width=\linewidth]{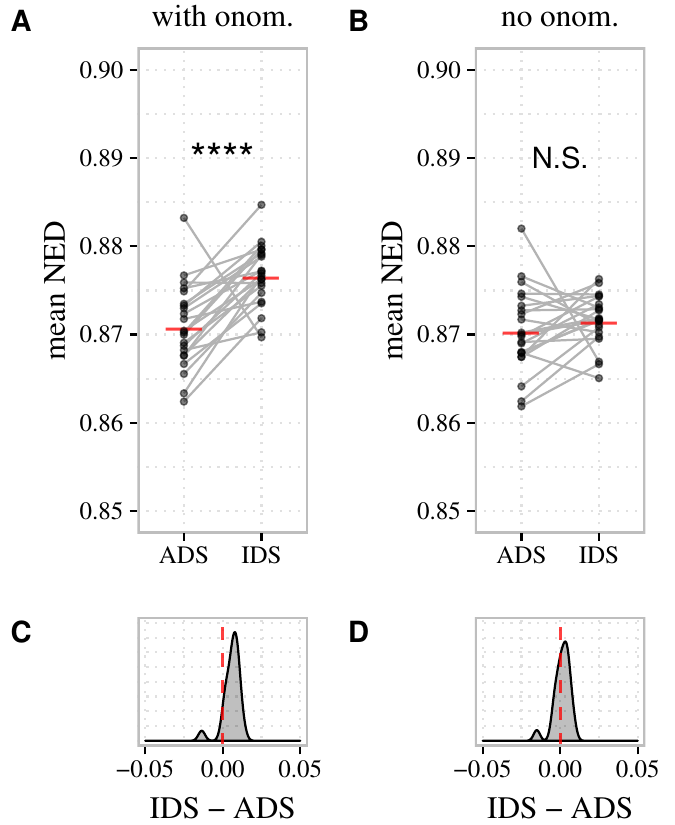}
    \hspace{0.5cm}
    
    \caption{Global phonological density scores (mean NED) for ADS and IDS, computed on lexicons matched for number of types across the two registers. Upper panels display the distribution of the scores across individual speakers, as well as means within a speech register (red horizontal lines). Grey lines connect data points corresponding to the same caregiver in both registers. Bottom panels show the distribution of IDS minus ADS score differences. Densities to the right of the red zero line denote higher scores for IDS. 
    A, C: Samples from base corpus.
    B, D: Samples from base corpus after onomatopoeia removal.
    $N.S.$: Non-significant difference. ****: $p < .0001$}
    \label{fig:exp2}
\end{figure}

The distribution of the difference in mean NEDs for IDS and ADS across parents is shown on panels A and C of Fig. \ref{fig:exp2}. Individual scores can be found in the Appendix Table A2. A pair-wise Student's \emph{t}-test showed a systematic pattern of larger normalized edit distances in IDS than ADS (IDS: $0.877$ \textit{vs.} ADS: $0.871$; $t(21) = 5.00$; $p < .0001$; \textit{Cohen's d} = $1.38$). This difference shows that, overall, the IDS lexicon contains words that are phonologically more distinctive than those in the ADS lexicon. In hindsight, a difference of this sort may have been expected as IDS has been found to contain ``babytalk'' or infant-directed vocabulary, \textit{i.e.}, a special vocabulary which includes onomatopoeias and phonological reduplications \cite{ferguson1964, fernald1993}. This hypothesis was verified in our dataset; we found that onomatopoeias and mimetic words (hereafter referred to solely as ``onomatopoeias'') constituted approximately 30\% of the average sample of IDS word types used in this experiment, whereas they represented less than 2\% of an average ADS sample (\textit{cf.} Appendix Table A2), this latter frequency being consistent with the use of mimetic words in Japanese observed in previous work \cite{saji2013}. 

In order to study the effect of onomatopoeias on phonological discriminability, we performed a post-hoc analysis by re-sampling words after removing all onomatopoeias from the base corpus. We then re-computed the mean NED for ADS and IDS. Individual scores can be found in the right side of the Appendix Table A2. A paired Student's \textit{t}-test revealed that the previously noted difference between IDS and ADS mean NED scores was no longer significant after onomatopoeia removal (IDS: $0.872$ \textit{vs.} ADS: $0.870$; $t(21) = 1.14; p > 0.05$; \textit{Cohen's d} = $0.31$, visual representation on panels B and D of Fig. \ref{fig:exp2}). Therefore, the IDS lexicon was found to be globally sparser than the ADS lexicon, and this effect seems to be principally driven by the unequal presence of onomatopoetic sounds in both speech registers.

Infant-directed words may facilitate lexical development not only by decreasing the overall phonological density of the lexicon, which directly impacts the clustering subproblem detailed in the introduction, but also in virtue of other intrinsic learning properties that would be relevant in a more complete model of early word learning. In the introduction we focused on the three key word learning subproblems of segmentation, word clustering, and phonetic categorization. At this point, it is imperative to point out that there are other factors that impact word learning in infancy above and beyond these particular processes. 

When asked about vocabulary specifically used when addressing infants, Japanese women report a set of words of which 40\% of the items are sound-symbolic \cite{mazuka2008}. An iconic relationship between an acoustic form and the semantics of the referent \cite{imai2014} has been shown to help 14-months old infants finding a word's referent \cite{miyazaki2013}, and it also facilitates the identification by pre-school children of the specific features of an action a verbal word form is referring to \cite{imai2008, kantartzis2011}. Additionally, around 65\% of the reported items contain reduplication of phonological patterns \cite{mazuka2008}, which may impact learning at a range of levels. 
Repetitive patterns may be more salient and generalizable than other equally complex patterns \cite{endress2007, endress2009}, and this salience could facilitate lexical acquisition in infants. This is supported by recent data showing that 9-month old English-learning infants segment words containing reduplications (e.g. \textit{neenee}) from running speech more easily than words without reduplications (e.g. \textit{neefoo}) \cite{ota2017}. Furthermore, English-learning 18-month-old infants appear to better learn novel object labels when these contain reduplications \cite{ota2016}. In fact, reduplication has been found to be a characteristic shared by many items from the specialized set of ``babytalk'' words in various languages \cite{ferguson1964}, in spite of the tendency to avoid such repetitive patterns in adult language \cite{leben1973}. 

Similarly to what was observed in the survey by \citeA{mazuka2008}, the majority of the word types tagged as onomatopoeias in our IDS corpus (i.e. around 30\% of the types) present reduplication and/or sound symbolism (\textit{e.g.,} \begin{CJK}{UTF8}{goth}わんわん\end{CJK} \textipa{/wa\textscn wa\textscn/} \textit{dog};  \begin{CJK}{UTF8}{goth}ころころ\end{CJK} \textipa{/koRokoRo/} \textit{light object rolling repeatedly}). Since infants seem to have a learning bias for words with these phonological characteristics, the higher proportion of onomatopoeias in IDS compared to ADS may provide an additional anchor for infant word learning. 

As a reviewer pointed out, it may seem counterintuitive at first to focus on the enhanced learnability of IDS-specific words, since children are expected to eventually master all words, whether they are specific to IDS or present in both IDS and ADS.  
However, we are not concerned here with all of language acquisition, but only with the possibility that top-down cues affecting sound category learning are more helpful in IDS compared to ADS. Thus, even if the words that are learned are not part of a general target lexicon, they might nonetheless present an easier word clustering subproblem, and in that way lead to a lexicon that can be used as seed for subsequent sound category extraction routines. 

In sum, we have found that IDS contains a higher proportion of onomatopoeias and mimetic words than ADS. Aside from their remarkable distinctiveness and salience, these items seem to contribute to decreasing the global density of the IDS lexicon. While words in IDS seem to be more spread in phonological space than words in ADS, phoneme-like representations may not yet be available to infants until a larger vocabulary is amassed \cite{lindblom1992, metsala1998, pierrehumbert2003, beckman2007}. As such, one may wonder if, similarly, words may be more distant in the acoustic space when taking the structural differences into account. Indeed, we notice that the effect size is almost twice as large for the phonological NED (\textit{Cohen's d} = $-1.38$) than for the acoustic discriminability (\textit{Cohen's d} = $-0.84$). However, given that they are not based on exactly the same tokens, it remains possible that the phonological advantage does not compensate for the acoustic disadvantage. Indeed, the difference in mean NED between IDS and ADS, while statistically significant, is numerically very small, representing a difference of less than one percent of a word. 
The following experiment examines the question of the effect of phonological structure on acoustic discriminability, by integrating both factors in one global discriminability measure.

\section{Experiment 3: Net Discriminability}

In Experiment 1, we found that when we looked at the exact same word types in both registers, the IDS tokens were acoustically more confusable than the ADS tokens, due to the increased variability of IDS word categories in the acoustic space. In other words, when removing the influence of structural peculiarities of the lexicons, IDS does not present an advantage over ADS in acoustic discriminability.
We then saw in Experiment 2 that the lexicons of IDS and ADS differed structurally. Words from the IDS lexicon were phonologically more distinct than those in the ADS lexicon, in part due to onomatopoeias and mimetic words.

Here, we put these two previous results together and ask the following question: When accounting for register-specific lexical structure, is the IDS lexicon acoustically clearer than the ADS lexicon? In other words, if we take a random pair of word tokens from two different word types found in the IDS recordings, are these tokens more or less acoustically distinct than a like-built pair in the ADS recordings? 

\subsection{Method}
\subsubsection{Sampling}
In order to observe the combined effects of the differences in phonological structure on acoustic discriminability, the same sampled lexicons used for Experiment 2 were used for this section, i.e. 100 lexicon subsets per register per speaker, matched in number of word types across speech registers. 

As it was done in Experiment 1, number of tokens per type were not matched in order to maximize the reliability of the ABX metric. Individual number of types can be seen in Table A2 of the Appendix, with total number of tokens shown in the Appendix Fig. A1.

\subsubsection{Computing acoustic discriminability}
Acoustic discriminability was computed as described in Experiment 1. A mean ABX score was computed per sampled lexicon subset. 
ABX scores were collapsed by computing the mean ABX score per speaker per register. 

\subsection{Results and Discussion}

\begin{figure}[!ht]
\centering
    \includegraphics[page=4, width=\linewidth]{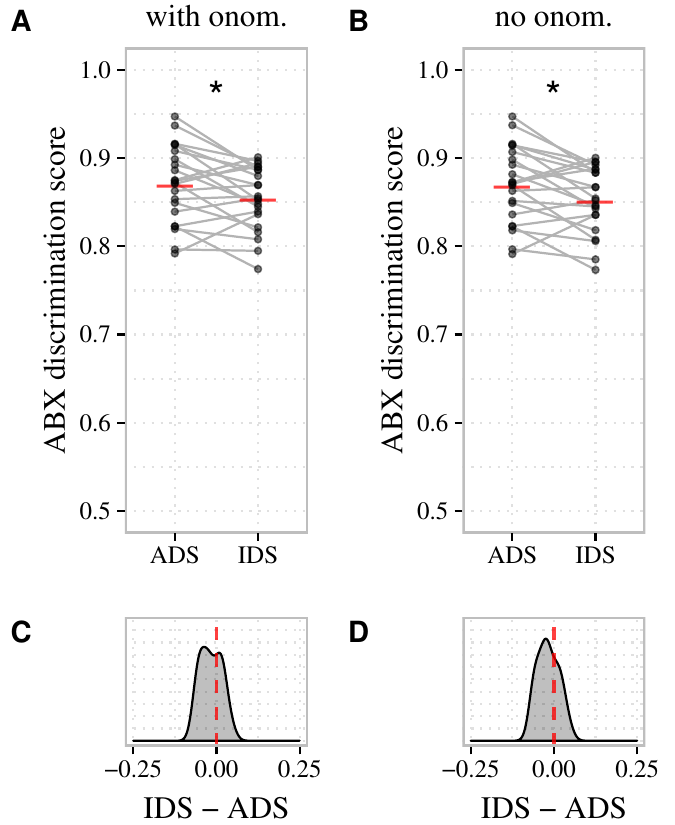}
    \hspace{0.5cm}
    
    \caption{Acoustic-based ABX word discrimination error in ADS and IDS computed on lexicons matched for number of word types across the two registers. Upper panels display the distribution of the scores across speakers, as well as means within a speech register (red horizontal lines). Grey lines connect data points corresponding to the same caregiver in both registers. Bottom panels show the distribution of IDS minus ADS score differences. Densities to the right of the red zero line denote higher error rates for IDS. 
    A, C: Samples from base corpus.
    B, D: Samples from base corpus after onomatopoeia removal.
    *: $p < .05$}
    \label{fig:exp3}
\end{figure}

We compared the mean ABX scores for ADS and IDS obtained on the sampled lexicons used in Experiment 2 (Fig. \ref{fig:exp3}). Individual scores can be found in the Appendix Table A2. A paired Student's \textit{t}-test revealed that mean $ABX_{score}$ were significantly larger for IDS than for ADS, whether onomatopoeias were included in the lexicon subsets ($ABX_{score}$ IDS: $86\%$ \textit{vs.} ADS: $87\%$; $t(21) = -2.37, p<0.05$; \textit{Cohen's d} = $-0.41$) or not ($ABX_{score}$ IDS: $85\%$ \textit{vs.} ADS: $87\%$; $t(21) = -2.57, p<0.05$; \textit{Cohen's d} = $-0.43$). As such, similar to what was found in Experiment 1, words are less discriminable in IDS than in ADS, even after taking into account the phonological specificities of the infant-directed lexicon. 

This result underlines the importance of assessing effects of language acquisition enhancers not only in terms of their statistical significance across parents ($p$ values, Cohen's d), but also quantitatively, i.e., in terms of their numerical strength when combined together. To see this more clearly, we computed the increase or decrease in the score under study as a percentage relative to the ADS score taken as a baseline. 

In Experiment 1, the decrement in discriminability in IDS was 4\% relative to ADS, and this effect was robust across participants ($Cohen's~d = -0.84$). In Experiment 2, the increase in NED represented a numerically smaller effect of less than 1\% for IDS relative to ADS. This effect was actually even more robust across participants ($Cohen's~d = 1.38$). Interestingly, when the two effects are combined (Experiment 3), the outcome is not determined by which effect was more statistically robust across the participants, but by which one was numerically larger. Indeed, the outcome yields a numerically small (1\% relative) decrement in discriminability, which is also much weaker across participants ($Cohen's~d = -0.41$).

\section{General Discussion} 

\begin{figure*}[ht] 
\centering
    \vspace{0.5cm}
    \begin{overpic}[height=6.5cm]{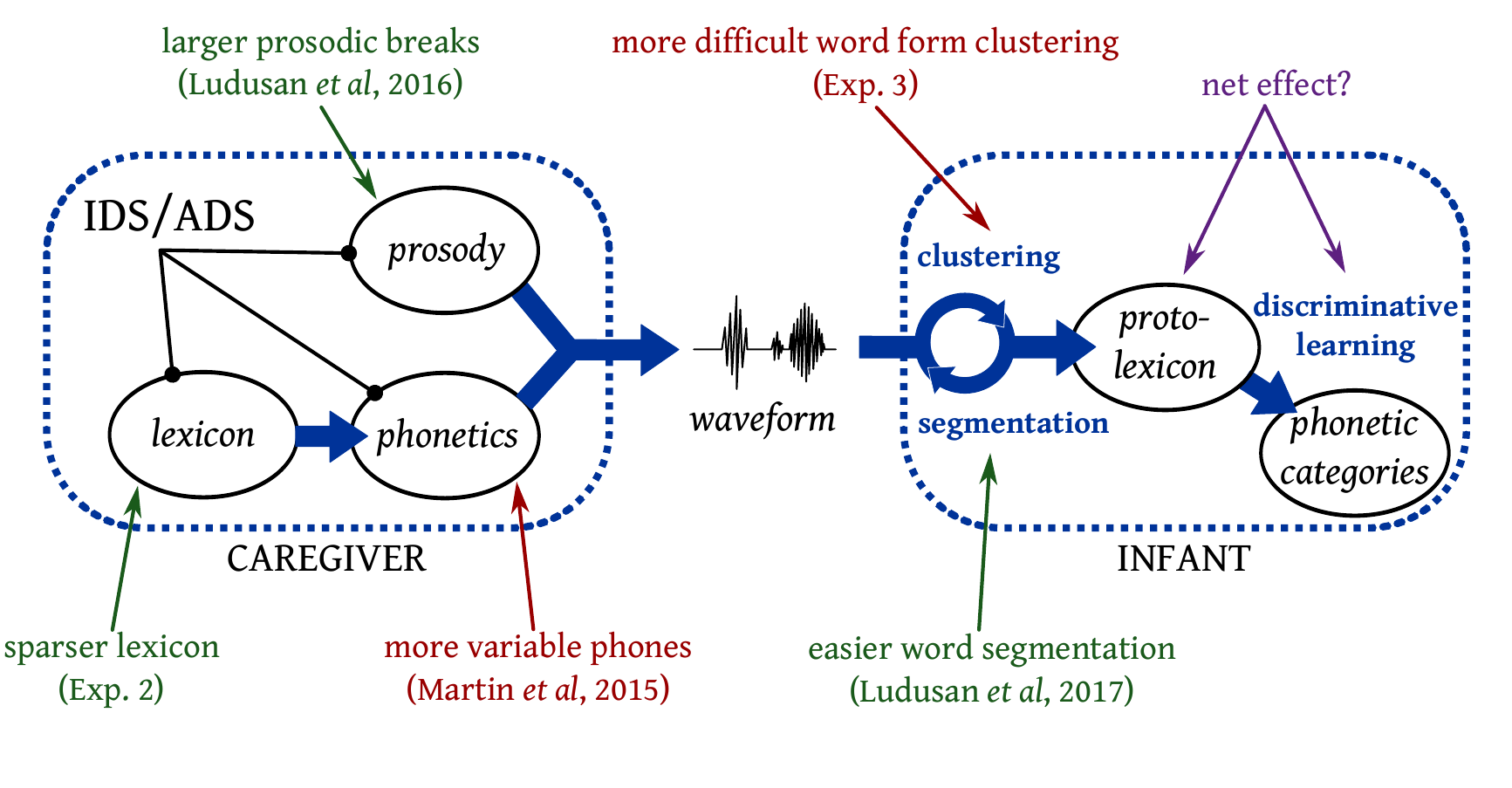}
        \hspace{0.5px}
    \end{overpic}
    \caption{Summary of IDS characteristics relative to ADS in a top-down model of phonetic category learning for the RIKEN corpus. Enhanced characteristics of IDS relative to ADS are shown in green, while those for which the opposite trend is observed are shown in red.} 
    \label{fig:schema_summary}
\end{figure*}


The Hyper Learnability Hypothesis (HLH) states that when talking to their infants, parents modify the linguistic properties of their speech in order to facilitate the learning process. In this paper, we focused on the learning of phonetic categories and reviewed two classes of theories in order to quantitatively assess the HLH: (1) bottom-up theories assume that phonetic categories emerge through the unsupervised clustering of acoustic information, (2) top-down theories assume that phonetic categories emerge through contrastive feedback from learned word types. Previous work has already addressed bottom-up theories: \citeA{martin2015} examined phonemes in a corpus of Japanese laboratory recordings and found that phonemes produced by caregivers addressing their 18- to 24-month old infants were less discriminable than ADS phonemes. This rules out the HLH for that corpus and bottom-up theories. In the present study, we focused on top-down theories using the same corpus, and investigated the acoustic discriminability of word types.

In Experiment 1, we compared the acoustic discriminability of words that are common to both speech registers, and found that words are \textit{less} discriminable in IDS than in ADS (an absolute decrease in $ABX_{score}$ of 4\%), likely because of increased within-category variability. This result parallels the increase in phonetic variability found in previous studies \cite{kirchhoff2005,cristia2014,mcmurray2013}, and it is consistent with the decreased phoneme discriminability measured by \citeA{martin2015}
. It is not consistent, however, with the claim that words in IDS are uttered in a more canonical way than in ADS (\citeNP{dilley_2014}, but see \citeNP{fais_2010,lahey_2014}). In Experiment 2, we turned to the structure of the phonological lexicon. We found that the IDS lexicon was globally more spread out than that of ADS, as shown by a larger normalized edit distance between words for the former. Interestingly, this effect was attributable mostly to a higher prevalence of onomatopoeias and mimetic words in IDS. These words have idiosyncratic phonological properties, such as reduplications, which are likely responsible for the increase of global distinctiveness found in the IDS lexicon, compared to the ADS lexicon. In Experiment 3, a final analysis measured the net effect of the opposite trends found in Experiments 1 and 2, and found that, on average, words were still less acoustically discriminable in IDS than in ADS, although the effect was now considerably reduced (an absolute decrease in $ABX_{score}$ of 1\%).


Overall, then, the word form clustering subproblem is not easier to solve by using IDS input than with ADS input; quite to the contrary, there is a numerically small but consistent trend in the opposite direction. Does this undermine the HLH for top-down theories of phonetic learning as a whole? Clearly, the answer is ``no'', since - as explained in the Introduction - HLH actually encompasses two other learning subproblems. 
We discuss relevant evidence on IDS-ADS differences bearing on each subproblem in turn.

Regarding the problem of finding word token boundaries, Ludusan and colleagues have started studying word form segmentation using either raw acoustics or text-like phonological representations as input. 
\citeA{ludusan2015} studied the performance of acoustic word form discovery systems on a corpus of American English addressed to 4- or 11-month-olds versus adults. 
The overall results are similar to those of Experiment 3, i.e., the two registers give similar outcomes, if anything, with a very small difference in favor of ADS, rather than the expected IDS. 
Computational models of word segmentation from running speech represented via acoustics are, however, well-known to underperform compared to models that represent speech via textual representations \cite{versteegh_2016}. Thus, in \citeNP{ludusan2017ACL}, we studied word form segmentation from text-like representations using the same RIKEN corpus as input, and a selection of state-of-the-art cognitively-based models of infant word segmentation. Results  showed an advantage of IDS over ADS for most algorithms and settings.

Beyond the question of whether segmentation is easier in IDS versus ADS, we cannot move on to the next learning subproblem without pointing out that, for future work to assess the net effect of register on word segmentation, one would need to know more about the size and composition of infants' early lexicon. In fact,  most accounts propose that the phonological system is extracted from the long-term lexicon, rather than on the fly from experience with the running spoken input (discussed in \citeNP{bergmann2017top}). In the present paper, we have done a systematic study of word discriminability across the whole set of words present in the corpus, as if infants could segment  the corpus exactly as adults do. This is, of course, unlikely. In fact, recent evidence suggests  that infants may be using a suboptimal segmentation algorithm  \cite{larsen2017}, which leads them to accumulate a ``protolexicon'' containing not only words, but also over- or under-segmented tokens that do not belong to the adult-like lexicon \cite{ngon2013}. Such protowords can nonetheless help with contrastive learning \cite{martin2013,fourtassi2014}. 

Regarding contrastive learning of phonetic categories, it is too early to know whether the net effect of register will be beneficial or detrimental. For instance, a detrimental effect of phonetic variability in a bottom-up setting can become beneficial in a top-down setting, by presenting infants with more varied input, and therefore preparing them for future between-speaker variability. This is illustrated in the supervised learning of phonetic categories in adults \cite{lively1993}. However, as suggested by \citeA{rost2010}, variability should be limited to acoustic cues that are not relevant to phonetic contrasts in order to promote learning.  
In order to fully assess the net effect of register, two important elements have to be clarified. First, one would need to have a fully specified model of contrastive learning itself. Candidate computational models have been proposed (e.g., \citeNP{feldman_learning_2009,fourtassi2014}), but not fully validated with realistic infant-directed speech corpora (but see \citeNP{versteegh_2016} for an application to ADS corpora). 



Throughout the above discussion, an important take-home message is that it is essential to posit well-defined, testable theories of infant learning, which can be evaluated using quantitative measures, even when fully specified computational models are not yet available. Individual studies focus only on few pieces of the puzzle and the magnitude of each evaluated effect must be observed relative to other effects. For instance, in our study, even the relatively large effect of IDS versus ADS on the discriminability of word forms found in Experiment 1 has to be compared to the much larger effect (by a factor of 2) of read versus spontaneous speech found within the ADS register. What we propose as a methodology is to break down theories of language acquisition into component parts, and to derive proxy measures for each component to derive a more systematic grasp of the quantitative effects of register. Before closing, we would like to discuss two limitations of the current study, one regarding the corpus and the other regarding the theory tested (the HLH).


The main limitation of the RIKEN corpus is that it was recorded in the laboratory and did not include naturalistic interactions between adults as they may occur in the home enviroment. The presence of an experimenter and props (toys, etc) in the laboratory setting may induce some degree of non-naturalness in the interaction, both with the infant, and with the adult.
\citeA{johnson2013} found that in Dutch, ADS is not a homogeneous register, and that it bears similarities with IDS when the addressed adult is familiar as opposed to unfamiliar.\footnote{In addition to these effects, Japanese and many other languages have a set of specialized morphemes that depend on familiarity between the talkers; this could have artificially increased the difference between IDS and ADS in the present corpus.} It remains to be assessed whether similar results are obtained in more ecological and representative IDS and ADS samples. In addition, the current study is limited by the relatively small size of the corpus. Because we analyzed each parent separately, the size of the analyzed lexicons was between 82 and 260 words, which may underrepresent the range of words heard in a home setting. Finally, our analysis is limited to Japanese. There is evidence that vowel hyperarticulation varies across languages \cite{benders2013,englund2005,kuhl1997}, and more generally that the specifics of the IDS register varies across culture (e.g., \citeNP{fernald1993, igarashi2013}).  It would therefore be important to replicate our methods in more ecological, cross-linguistic corpora. Fortunately, the availability of wearable recording systems such as the LENA$\copyright $ device \cite{greenwood2011} increases the prospects of automatizing the collection and analysis of  naturalistic speech \cite{soderstrom2013}.

The second limitation of the current study is that we restricted our quantitative analysis to the testing of the HLH. However, the HLH is not the only hypothesis that can be addressed. Other theories have been proposed regarding the etiology and role of IDS in the linguistic development of infants (i.e., why caregivers use it, and what are the actual effects on the child). Some modifications of the input may indeed have pedagogical functions (enhancing learnability), while other modifications may decrease learnability while increasing some other factor in the parent-infant interaction.    
For instance, it has been documented that mothers sometimes violate the grammar of their language when teaching new words, probably in order to place the novel word in a sentence-final position \cite{aslin1996}, which is salient because of properties of short-term memory. Similarly, it has sometimes been suggested that caregivers inadvertently sacrifice phonetic precision in order to make infants more comfortable and/or more receptive to the input \cite{papousek1991,reilly1996}. Increased phonetic variability in IDS at the phonemic level may stem from a slower speaking rate \cite{mcmurray2013}, or from exaggerated prosodic variations \cite{fernald1989, soderstrom2007,martin2016}, or possibly from gestural modifications that convey positive affect, such as smiling  \cite{benders2013}, increased breathiness \cite{miyazawa2017} or even a vocal tract that is shortened to resemble the child's own \cite{kalashnikova2017}. 
According to a study by \citeA{trueswell2016}, successful word learning interactions tend to be those in which actions performed by both caregivers and infants are precisely synchronised, with time-locking of gaze, speech and gestures. By focusing on efficiently capturing the infant's attention, caregivers could create an optimal learning environment, in spite of potential degradations brought upon lexical acoustic clarity. A similar interpretation is held by authors such as \citeA{csibra2006}, who argue that one of the main roles of IDS is to inform the infant that speech is being directed to her, thus highlighting the pedagogical nature of the interaction as a whole. In this view, the goal of caregivers would not be to provide clearer input, but to make language interactions and their attached learning situations more exciting and attractive to infants. 

Another direction entirely, is to propose that IDS may help infants to \textit{produce} language. \citeA{ferguson1964} describes ``babytalk'' as a subset of phonologically-simplified words due to reduced consonant clusters, use of coronals instead of velars, word shortening, etc. These adaptations would make it easier for developing infants to imitate the words, and/or they may be inspired by previous generations' production errors. In fact, previous work performed on our corpus shows that, if anything, the structural properties of words in our IDS sample better fit early patterns of Japanese infant speech production than those of words in ADS \cite{tsuji2014segmental}.
While the causal relationship between babytalk use and infant word production should be further assessed experimentally, the phonological properties of our IDS corpus suggest that, to some extent, parental input may be encouraging infant word production.
  
In brief, while the HLH focuses on the change in informational content of IDS which may boost (or hinder) the learnability of particular linguistic structures, IDS could have a beneficial effect on completely different grounds: enhancing overall attention or positive emotions which would increase depth of processing and retention, or facilitating production, thereby counteracting the inadvertent acoustic degradation of local units of speech such as words and phonemes. For these alternative theories of HLH to be testable within our quantitative approach, we would need to formulate these theories with enough precision that they can either be implemented, or proxies can be derived to analyze realistic corpora of caregivers/infants interactions.

To conclude, the last 50 years we have learned a great deal about how IDS and ADS differ, yet much remains to be understood. We believe it is crucial in this quest to bear in mind a detailed model of early language acquisition, and to submit predictions of this model to systematic, quantitative tests. 

\section{Author contributions} 

R. Mazuka oversaw the collection and coding of the corpus. A. Martin wrote the algorithms for extracting words and their phonological structure. R. Thiollière provided coding support with the ABX task. A. Cristia directed the literature review. B. Ludusan assisted with preparation of the ADS-RS comparison. A. Guevara-Rukoz and E. Dupoux carried out the acoustical and phonological analyses and, along with A. Cristia, produced the first draft. All authors contributed to 
the writing of this manuscript.

\section{Acknowledgements}  

This work was supported by the European Research Council (Grant ERC-2011-AdG-295810 BOOTPHON), the Agence Nationale pour la Recherche (Grants ANR-2010-BLAN-1901-1 BOOTLANG, ANR-14-CE30-0003 MechELex, ANR-10-IDEX-0001-02 PSL*, and ANR-10-LABX-0087 IEC), the Fondation de France, the Japan Society for the Promotion of Science (Kakenhi Grant 24520446, to A. Martin), and the Canon
Foundation in Europe. 
We thank Bob McMurray and two anonymous reviewers for helpful feedback.

\setcounter{secnumdepth}{0}
\bibliography{idslex}

\clearpage

\includepdf[pages={-}]{appendix_jou.pdf} 

\end{document}